%% file: icml10arxiv.tex
\documentclass{article}

\usepackage{myCommands}

\usepackage[articlestyle]{mpemath}
\usepackage{notation}

\usepackage{graphicx}
\usepackage{subfigure}

\usepackage{natbib}

\newcommand{\trchoose}[2]{#2}

\usepackage{hyperref}

\setitemize{nolistsep}
\setenumerate{nolistsep}
\setdescription{nolistsep}

\usepackage[accepted]{icml2010}

\icmltitlerunning{Feature Selection Using Regularization}

\newcommand{\lbl}{\label}
\newcommand{\lblref}[1]{\tag{\ref{#1}}}

\newcommand{\defthm}[3]{
\begin{thm}
#3\label{#2}
\end{thm}
\trchoose{Because of the technical nature of the proof, it is omitted and can be found in \cite{techrep}.}{Because of the technical nature of the proof, it is deferred to
the appendix.}
\newcommand{#1}{
\renewcommand{\lbl}{\lblref}
\par \phantomsection \label{#2_proof}
\noindent{\bf Theorem \ref{#2}}.
\emph{ #3 }
\renewcommand{\lbl}{\label}
} \par }

\newcommand{\defprop}[3]{
\begin{prop}
#3\label{#2}
\end{prop}
\trchoose{Because of the technical nature of the proof, it is omitted and can be found in \cite{techrep}.}{Because of the technical nature of the proof, it is deferred to
the appendix.}
\newcommand{#1}{
\renewcommand{\lbl}{\lblref}
\par \phantomsection \label{#2_proof}
\noindent{\bf Theorem \ref{#2}}.
\emph{ #3 }
\renewcommand{\lbl}{\label}
} \par }

\begin{document}

\twocolumn[
\icmltitle{Feature Selection Using Regularization in Approximate Linear
Programs for Markov Decision Processes}

\icmlauthor{Marek Petrik*}{petrik@cs.umass.edu}
\icmlauthor{Gavin Taylor$\dagger$}{gvtaylor@cs.duke.edu}
\icmlauthor{Ron Parr$\dagger$}{parr@cs.duke.edu}
\icmlauthor{Shlomo Zilberstein*}{shlomo@cs.umass.edu}
\icmladdress{* Department of Computer Science,
            University of Massachusetts,
            Amherst, MA 01003 USA}
\vspace{-0.1in}
\icmladdress{$\dagger$ Department of Computer Science,
            Duke University, Durham, NC 27708 USA}


\icmlkeywords{approximate linear programming, regularization, lasso}
\vskip 0.1in
]

\input{intro.tex}
\input{notation.tex}
\input{alpbackground.tex}
\input{ralp.tex}
\input{sampling.tex}

\input{experimental.tex}
\input{conclusion.tex}

\bibliography{icml10}
\bibliographystyle{icml2010}

\trchoose{}{
\newpage
\appendix
\input{proofs.tex}

\input{homotopy.tex}}

\end{document}

%% file: intro.tex
\begin{abstract}
Approximate dynamic programming has been used successfully in a large variety
of domains, but it relies on a small set of provided approximation features to
calculate solutions reliably. Large and rich sets of features can cause
existing algorithms to overfit because of a limited number of samples. We
address this shortcoming using $L_1$ regularization in approximate linear
programming. Because the proposed method can automatically select the
appropriate richness of features, its performance does not degrade with an
increasing number of features. These results rely on new and stronger sampling
bounds for regularized approximate linear programs.  We also propose a
computationally efficient homotopy method. The empirical evaluation of the
approach shows that the proposed method performs well on simple MDPs and
standard benchmark problems.
\end{abstract}

\section{Introduction}

Solving large Markov Decision Processes (MDPs) is a very useful, but
computationally challenging problem addressed widely by reinforcement
learning.  It is widely accepted that large MDPs can only be solved
approximately.  This approximation is commonly done by relying on linear value
function approximation, in which the value function is chosen from a
small-dimensional vector space of features.  While this framework offers
computational benefits and protection from the overfitting in the training
data, selecting an effective, small set of features is difficult and requires
a deep understanding of the domain. Feature selection, therefore, seeks to
automate this process in a way that may preserve the computational simplicity
of linear approximation~\cite{parr07,mahadevan08}. We show in this paper that
$\lone$-regularized approximate linear programs (RALP) can be used with very
rich feature spaces.

RALP relies, like other value function approximation methods, on samples of
the state space. The value function error on states that are not sampled is
known as the \emph{sampling error}. This paper shows that regularization in
RALP can guarantee small sampling error.  The bounds on the sampling error
require somewhat limiting assumptions on the structure of the MDPs, as any
guarantees must, but this framework can be used to derive tighter bounds for
specific problems in the future. The relatively simple bounds can be used to
determine automatically the regularization coefficient to balance the
expressivity of the features with the sampling error.

We derive the approach with the $\lone$ norm, but it could be used with other
regularizations with small modifications. The $\lone$ norm is advantageous for
two main reasons. First, the $\lone$ norm encourages the sparse 
solutions, which can  reduce the computational requirements. Second, the
$\lone$ norm preserves the linearity of RALPs; the $L_2$ norm would require
quadratic optimization.

Regularization using the $\lone$ norm has been widely used in regression
problems by methods such as LASSO~\cite{tibshirani96} and Dantzig
selector~\cite{Candes2007}. The value-function approximation setting is,
however, quite different and the regression methods are not \emph{directly}
applicable. Regularization has been previously used in value function
approximation~\cite{icml09,farahmand08,kolter09a}. In comparison with
LARS-TD~\cite{kolter09a}, an $L_1$ regularized value function approximation
method, we explicitly show the influence of regularization on the sampling
error, provide a well-founded method for selecting the regularization
parameter, and solve the full control problem. In comparison with existing
sampling bounds for ALP~\cite{farias2001}, we do not assume that the optimal
policy is available, make more general assumptions, and derive bounds that are
independent of the number of features.

Our approach is based on approximate linear programming~(ALP), which offers
stronger theoretical guarantees than some other value function approximation
algorithms. We describe ALP in \aref{sec:alpbackground} and RALP and its basic
properties in \aref{sec:ralp}. RALP, unlike ordinary ALPs, is guaranteed to
compute bounded solutions. We also briefly describe a homotopy algorithm for
solving RALP, which exhibits anytime behavior by gradually increasing the norm
of feature weights. To develop methods that automatically select features with
generalization guarantees, we propose general sampling bounds in
\aref{sec:sampling}. These sampling bounds, coupled with the homotopy method,
can automatically choose the complexity of the features to minimize
over-fitting. Our experimental results in \aref{sec:experimental} show that
the proposed approach with large feature sets is competitive with LSPI
when performed even with small feature spaces hand selected for standard
benchmark problems.  \aref{sec:conclusion} concludes with future work and a
more detailed relationship with other methods.

%% file: notation.tex
\section{Framework and Notation} \label{sec:notation}

In this section, we formally define Markov decision processes and linear value
function approximation.  A \emph{Markov Decision Process} is a tuple
$(\states,\actions,P,r,\gamma)$, where $\states$ is the possibly infinite set
of states, and $\actions$ is the finite set of actions. $P: \states \times
\states \times \actions \mapsto [0,1]$ is the transition function, where
$P(s',s,a)$ represents the probability of transiting to state $s'$ from state
$s$, given action $a$. The function $r: \states \times \actions \mapsto \Real$
is the reward function, and $\gamma$ is the discount factor.  $P_a$ and $r_a$
are used to denote the probabilistic transition matrix and reward vector for
action $a$.

We are concerned with finding a value function $v$ that maps each state
$s\in\states$ to the expected total $\gamma$-discounted reward for the
process. Value functions can be useful in creating or analyzing a policy
$\pol: \states \times \actions \rightarrow [0,1]$ such that for all
$s\in\states$, $\sum_{a\in\actions} \pol(s,a) = 1$.  The
transition and reward functions for a given policy are denoted by $P_\pol$ and
$r_\pol$.  The value function update for a policy $\pol$ is denoted by
$\Bell_\pol$, and the Bellman operator is denoted by $\Bell$. That is:
\begin{align*} \Bell_\pol v &= \disc P_\pol v + r_\pol & \Bell v &=
\max_{\pol\in\policies} \Bell_\pol v.  \end{align*} The optimal value function
$v^*$ satisfies $\Bell v^* = v^*$.

We focus on \emph{linear value function approximation} for discounted
infinite-horizon problems, in which the value function is represented as a
linear combination of \emph{nonlinear basis functions (vectors)}. For each
state $s$, we define a vector  $\phi(s)$ of features. The rows of the basis
matrix $\repm$ correspond to $\phi(s)$, and the approximation space is
generated by the columns of the matrix. That is, the basis matrix $\repm$, and
the value function $v$ are represented as:
\[
\repm = \begin{pmatrix} \mbox{ ---} & \phi(s_1)\tr & \mbox{--- } \\  &
\vdots & \end{pmatrix} \qquad v = \repm w .
\]
This form of linear representation allows for the calculation of an
approximate value function in a lower-dimensional space, which provides
significant computational benefits over using a complete basis; if the number
of features is small, this framework can also guard against overfitting noise
in the samples.

\begin{defn} \label{def:representable}
A value function, $v$, is \emph{representable} if $\val \in \rep \subseteq
\Real^{|\states|}$, where $\rep = \cspan{\repm}$. The set of
$\epsilon$-\emph{transitive-feasible} value functions is defined for $\epsilon
\geq 0$ as follows: $\tf(\epsilon) = \{ \val \in \Real^{|\states|} \ss \val \geq \Bell \val - \epsilon\one$. Here $\one$ is a vector of all ones. A value function is
\emph{transitive-feasible} when $v \geq \Bell v$ and the set of transitive-feasible functions is defined as $\tf = \tf(0)$.
\end{defn}
Notice that the optimal value function $v^*$ is transitive-feasible, and that
$\rep$ is a linear space.


%% file: alpbackground.tex
\section{Approximate Linear Programming} \label{sec:alpbackground}

The approximate linear programming (ALP) framework is an approach for
calculating a value function approximation for large MDP with a set of
features $\Phi$ that define a linear space
$\rep$~\cite{schweitzer85,defarias03}. The ALP takes the following form:
\begin{mprog*}
\minimize{\val \in\rep} \sum_{s\in \states} \tobj(s) \val(s)
\stc r(s,a) + \gamma  \sum_{s' \in \states} P(s',s,a) \val(s') \leq \val(s)
\end{mprog*}
where $\tobj$ is a distribution over the initial states and the constraints
are for all $(s,a) \in (\states,\actions)$; that is $\sum_{s\in\states}
\tobj(s) = 1$.  To ensure feasibility, one of the features is assumed to be
constant. Therefore, in the remainder of the paper, we make the following
standard assumption~\cite{schweitzer85}, which can be satisfied by setting the
first column of $\rep$ to $\one$.
\begin{asm}\label{asm:contains_one}
For all $k \in \Real$, we have that $k \cdot \one \in \rep$, where $\one$ is a
vector of all ones.
\end{asm}

For simplicity and generality of notation, we use $\sBell$ to denote the ALP
constraint matrix, so $\sBell\val\leq\val$ is equal to the set of constraints
$\{\Bell_a \val \leq \val : \forall a\in\sA \}$. Then, we can rewrite the ALP
as follows:
\begin{mprog} \label{mpr:lp}
\minimize{\val} \tobj\tr \val
\stc \sBell \val \leq \val \quad \val \in \rep
\end{mprog}
Notice that the constraints in the ALP correspond to the definition of transitive-feasible functions in \aref{def:representable}. A succinct notation of the ALP constraints can then use the set of transitive-feasible functions as $\val \in \rep \cap \tf$.

The constraint $\val \in \rep$ implies that $\val = \repm w$ and therefore the
number of variables in \eqref{mpr:lp} corresponds to the number of features.
Typically, this is a small number. However, the number of required constraints
in ALP is $|\states| \times |\actions|$, which is oftentimes impractically
large or infinite. The standard solution is to sample a small set of
constraints according to a given distribution~\cite{defarias03}. It is then
possible to bound the probability of violating a randomly chosen constraint.
There are, however, a few difficulties with this approach. First, leaving
constraints out can lead to an unbounded linear program. Second, in practice
the distribution over the constraints can be very different from the
distribution assumed by the theory.  Finally, the bound provides no guarantees
on the solution quality.

ALP has often under-performed ADP methods in practice; this issue has been
recently studied and partially remedied~\cite{petrik09,Desai2009}. Because
these methods are independent of the proposed modifications, we only focus on
standard approximate linear programs.

We show next that RALP with sampled constraints not only guarantees that the
solution is bounded and provides worst-case error bounds on the value
function, but also is independent of the number of features. As a result, the
ALP formulation does not require a small number of features to be selected in
advance.


%% file: ralp.tex
\section{Regularized Approximate Linear Programming} \label{sec:ralp}

In this section, we introduce $\lone$-regularized ALP (RALP) as an approach to
automate feature selection and alleviate the need for all constraints in
standard ALP. Adding $\lone$ regularization to ALP permits the user to supply
an arbitrarily rich set of features without the risk of overfitting.

The RALP for basis $\Phi$ and $\lone$ constraint $\rc$ is defined as follows:
\begin{mprog} \label{mpr:alp_lone}
\minimize{\bw} \tobj\tr \Phi\bw
\stc \sBell\Phi\bw\ \leq \Phi\bw \quad \norm{\bw}_{1,\reg} \leq \rc,
\end{mprog}
where $\|w\|_{1,\reg} = \sum_{i} \reg(i) w(i)$. Note that RALP is a
generalization of ALP; when $\rc$ approaches infinity, the RALP solution
approaches the ALP solution. The objective value of \eqref{mpr:alp_lone} as a
function of $\rc$ is denoted as $\objregone(\rc)$.

We generally use $\reg = \one_{-1}$, which is a vector of all ones except the
first position, which is 0; because the first feature is the constant feature,
we do not include it in the regularization. The main reasons for excluding the
constant feature are that the policy is independent of the constant shifts,
and the homotopy method we propose requires that the linear program is easy to
solve when $\rc = 0$.

Alternatively, we can formulate RALP in \eqref{mpr:alp_lone} as a minor
modification of ALP in equation \eqref{mpr:lp}. This is by modifying $\rep$ to satisfy the $\lone$ norm as: 
\[\rep(\rc) = \{ \repm w \ss \| w \|_{1,\reg} \leq \rc \}.\]

Notice that RALP introduces an additional parameter $\rc$ over ALP.  As with
$\lone$ regularization for regression, this raises some concerns about a
method for choosing the regularization parameter.  Practical methods, such as
cross-validation may be used to address this issue.  We also propose an
automated method for choosing $\rc$ in \aref{sec:sampling} based on the
problem and sampling parameters.

%% file: sampling.tex
\section{Sampling Bounds} \label{sec:sampling}

The purpose of this section is to show that RALP offers two main benefits over
ALP.  First, even when the constraints are sampled and incomplete, it is
guaranteed to provide a feasible solution. Since feasibility does not imply
that the solution is close to optimal, we then show that under specific
assumptions --- such as smooth reward and transition functions --- RALP
guarantees that the error due to the missing constraints is small.

To bound the error from sampling, we must formally define the samples and how
they are used to construct ALPs. We consider the following two types of
samples $\tilde\samples$ and $\bar\samples$ defined as follows.  \begin{defn}
\emph{One-step simple samples} are defined as follows:
$\tilde{\samples} \subseteq \{ (s,a,(s_1 \ldots s_n),\rew(s,a)) \ss s \in
\states, \; a \in \actions \},$ where $s_1 \ldots s_n$ are selected i.i.d.
from the distribution $\tran(s,a,\cdot)$.
\emph{One-step samples with expectation} are defined as follows:
$ \bar{\samples} \subseteq \{ (s,a,\tran(s,a,\cdot),\rew(s,a)) \ss s \in \states, \; a \in \actions \}.$
\end{defn}
Often the samples only include state transitions, as $\tilde{\samples}$
defines. The more informative samples $\bar{\samples}$ include the probability
distribution of the states that follow the given state and action, as follows:
\begin{align}
\label{eq:upper_second_samples4}
\bar{\Bell}(\val)(\bar{s}) &= \rew(\bar{s}, a) + \disc \sum_{s' \in \states}
\tran(\bar{s},a,s')\val(s'),
\end{align}
where $(\bar{s},a,\tran(\bar{s},a,\cdot),\rew(\bar{s},a)) \in \bar{\samples}$.
The less-informative $\tilde\samples$ can be used as follows:
\begin{align}
\label{eq:upper_second_samples3}
\tilde{\Bell}(\val)(\tilde{s}) &= \rew(\tilde{s}, a) + \disc \frac{1}{n}
\sum_{i=1}^n v(s_i),
\end{align}
where $(\tilde{s},a,(s_1 \ldots s_n),\rew(\tilde{s},a))\in \tilde{\samples}$.
The corresponding transitive-feasible sets $\bar{\tf}$ and $\tilde{\tf}$ are
defined similarly. The ALPs can be constructed based on samples as
\aref{fig:sampled_alp} shows. \emph{Full ALP} corresponds to the RALP
formulation in \eqref{mpr:alp_lone}, when $\rep$ is constricted with $\lone$
regularization. In comparison, \emph{sampled ALP} is
missing some of the constraints while \emph{estimated ALP} is both missing
some constraints, and the included constraints may be estimated imprecisely.

\begin{figure*}
\centering
\begin{small}
\begin{minipage}{0.3\linewidth}
\centering
\textbf{Full ALP}
\[ \tobj = \frac{1}{|\states|} \sum_{s \in \states} \feats(s) \]
\begin{mprog} \label{mpr:alp_full}
\minimize{\val} \tobj\tr \val
\stc \val \in \tf \qquad  \val \in \rep(\rc)
\end{mprog}
\end{minipage}
\begin{minipage}{0.3\linewidth}
\centering
\textbf{Sampled ALP}
\[ \bar{\tobj} = \frac{1}{|\bar{\samples}|} \sum_{(s,\ldots) \in \bar{\samples}} \feats(s) \]
\begin{mprog} \label{mpr:alp_sampled}
\minimize{\val} \bar{\tobj}\tr \val
\stc \val \in \bar{\tf} \qquad \val \in \rep(\rc)
\end{mprog}
\end{minipage}
\begin{minipage}{0.3\linewidth}
\centering
\textbf{Estimated ALP}
\[ \bar{\tobj} = \frac{1}{|\tilde{\samples}|} \sum_{(s,\ldots) \in \tilde{\samples}} \feats(s) \]
\begin{mprog} \label{mpr:alp_estimated}
\minimize{\val} \bar{\tobj}\tr \val
\stc \val \in \tilde{\tf} \qquad \val \in \rep(\rc)
\end{mprog}
\end{minipage}
\end{small}
\caption{Constructing ALP From Samples} \label{fig:sampled_alp}
\end{figure*}

The following two assumptions quantify the behavior of the ALP with respect to missing and imprecise constraints respectively. The first assumption limits the error due to missing transitions in the sampled Bellman operator $\bar{\Bell}$.
\begin{asm}[Constraint Sampling Behavior] \label{asm:sampling_behavior}
The representable value functions satisfy that:
\[\tf \cap \rep(\rc) \subseteq \bar{\tf} \cap \rep(\rc) \subseteq \tf(\epsilon_p) \cap \rep(\rc), \]
and that for all representable value functions $\val \in \rep(\rc)$ we have that
$|(\tobj - \bar{\tobj})\tr \val | \leq \epsilon_c(\rc)$.
\end{asm}
The constant $\epsilon_p$ bounds the potential violation of the ALP
constraints on states that are not provided as a part of the sample. In
addition, all value functions that are transitive-feasible for the full
Bellman operator are transitive-feasible in the sampled version; the sampling
only removes constraints on the set. The constant $\epsilon_c$ essentially
represents the maximal error in estimating the objective value of ALP for any representable value function.

The next assumption quantifies the error on the estimation of the transitions
of the estimated Bellman operator $\tilde\Bell$.
\begin{asm}[Constraint Estimation Behavior] \label{asm:estimation_behavior}
The representable value functions satisfy that:
\[\bar{\tf}(-\epsilon_s) \cap \rep(\rc) \subseteq \tilde{\tf} \cap \rep(\rc) \subseteq \bar{\tf}(\epsilon_s) \cap \rep(\rc),\]
where $\bar\samples$ and $\tilde\samples$ (and therefore $\bar\tf$ and $\tilde\tf$) are defined for identical sets of states.
\end{asm}

These assumptions are quite generic in order to apply in a wide range of
scenarios. The main idea behind the assumptions is to bound by how much a
feasible solution in the sampled or estimated ALP can violate the true ALP
constraints. These assumptions may be easily satisfied, for example, by
making the following Lipschitz continuity assumptions.
\begin{asm} \label{asm:lipschitz}
Let $k:\states\rightarrow\Real^n$ be a map of the state-space to a normed
vector space. Then for all $x,y,z\in \sS$ and all features (columns) $\phi_i
\in\Phi$, we define $K_\rew$, $K_P$, and $K_\phi$ such that
\begin{align*}
|\rew(x)-\rew(y)| &\leq K_\rew \norm{k(x)-k(y)}\\
|p(z|x,a)-p(z|y,a)| &\leq K_P \norm{k(x)-k(y)}\\
|\phi_i(x)-\phi_i(y)| &\leq K_\phi \norm{k(x)-k(y)}
\end{align*}
\end{asm}
\defprop{\corlinearvalues}{cor:linear_values}{
Assume \aref{asm:lipschitz} and that for any $s\in\states$ there exists a state
$\bar{s}\in\bar{\samples}$ such that $\| \bar{s} - s \| \leq c$. Then
\aref{asm:sampling_behavior} and \aref{asm:estimation_behavior} hold with
$\epsilon_p(\rc) = c K_r + c \rc ( K_\phi + \disc K_P )$
}
The importance of this bound is that the violation on constraints that were not sampled grows linearly with the increasing coefficient $\rc$. As we show below, this fact can be used to determine the optimal value of $\rc$. For the
sake of brevity, we do not discuss the estimation error bounds $\epsilon_s$ in
more detail, which can be easily derived from existing
results~\cite{petrik09}.

We are now ready to state the following general bounds on the approximation error of a RALP.
\defthm{\thmregboundoffline}{thm:regbound_offline}{[Offline Error Bound]
Assume Assumptions \ref{asm:contains_one}, \ref{asm:sampling_behavior},
and \ref{asm:estimation_behavior}. Let $\hat \val$, $\bar \val$, $\tilde \val$
be the optimal solutions of \eqref{mpr:alp_full}, \eqref{mpr:alp_sampled}, and
\eqref{mpr:alp_estimated}, respectively (see \aref{fig:sampled_alp}). Let $
\epsilon = \frac{2}{1-\disc} \min_{\val\in\rep(\rc)} \| \val -
\val^*\|_\infty$ Then, the following inequalities hold:
\begin{align*}
\| \hat \val - \val^* \|_{1,\tobj} &\leq \epsilon  \\
\| \bar \val - \val^* \|_{1,\tobj} &\leq \epsilon + 2 \epsilon_c(\rc) + 2\frac{\epsilon_p(\rc)}{1-\disc} \\
\| \tilde \val - \val^* \|_{1,\tobj} &\leq \epsilon + 2 \epsilon_c(\rc) + \frac{3\epsilon_s(\rc) + 2\epsilon_p(\rc)}{1-\disc}
\end{align*}}
Notice that because the weight $\tobj$ is often chosen arbitrarily, the bounds
may be simply derived for $\|\cdot\|_{1,\bar{\tobj}}$. In that case,
$\epsilon_c = 0$.  Unlike most of the existing ALP bounds, we
focus on bounding the error of the value function, instead of bounding the
number of violated constraints.

Consider the implications of these bounds combined with
the Lipschitz assumptions of \aref{cor:linear_values}.  It is clear that
reducing $\rc$ tightens \aref{thm:regbound_offline}, but causes the set
$\rep(\rc)$ to shrink and become more restrictive; this suggests a tradeoff to
be considered when setting the regularization parameter. The bound also
illustrates the importance of covering the space with samples; as the distance
between the samples $c$ approaches zero, the bound tightens.  In short, the
Lipschitz assumptions limit how quickly the constraints can change between
sampled states.  As sampled states get closer or the reward, feature, and
probability functions become smoother, constraints between samples become more
and more restricted; however, smoother basis functions may mean a less
expressive space.  Similar tradeoffs are likely to appear however
\aref{asm:sampling_behavior} and \aref{asm:estimation_behavior} are fulfilled.

The offline error bounds in \aref{thm:regbound_offline} can be used to
guarantee the performance of a RALP for a fixed number of samples and the
regularization coefficient $\rc$. It does not, however, prescribe how to
choose the regularization coefficient for a given set of samples. To do that,
we have to derive bounds for an actual value function $\val$. When the samples
are known, these bounds are typically tighter than the offline error bound.
\defthm{\thmregboundonline}{thm:regbound_online}{[Online Error Bound]
Assume \aref{asm:contains_one} and let $\val \in \tilde{\tf} \cap \rep(\rc)$ be an arbitrary feasible solution of the estimated ALP \eqref{mpr:alp_estimated}. Then:
\[ \| \val^* - \val \|_{1,\tobj} \leq \bar{\tobj}\tr \val - \tobj\tr \val^* + \epsilon_c(\rc) + 2\frac{\epsilon_s(\rc) + \epsilon_p(\rc)}{1-\disc}. \]
}

\begin{figure}
\centering
\includegraphics[width=0.7\linewidth]{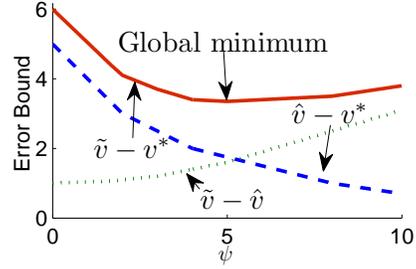}
\caption{Sketch of error \emph{bounds} as a function of the regularization
coefficient. Here, $\hat v$ is the value function of the full ALP, $\tilde v$
is the value function of the estimated ALP, and $v^*$ is the optimal value
function.} \label{fig:sampling_bounds}
\end{figure}

Here we briefly introduce a homotopy method that not only speeds the computation of the RALP solution, but also can be used to find the optimal value of $\rc$. Because the homotopy method only relies on standard linear programming analysis and is somewhat technical, the detailed description is provided in \trchoose{\cite{techrep}}{\aref{sec:Homotopy} in the appendix}.

The main idea of the homotopy method is to first calculate $\objregone(0)$ and then trace the optimal solution for increasing values of $\rc$. The optimal solution $w(\rc)$ of the RALP \eqref{mpr:alp_lone} is a piecewise linear function of $\rc$.  At any point in time, the algorithm keeps track of
a set of active -- or non-zero -- variables $w$ and a set of active constraints, which are satisfied with equality. In the linear segments, the number of active constraints and variables are identical, and the
non-linearity arises when variables and constraints become active or inactive. Therefore, the linear segments are traced until a variable becomes inactive or a constraint becomes active. Then, the dual solution is traced until a constraint becomes inactive or a variable becomes active.


Since the homotopy algorithm solves for the optimal value of the RALP
\eqref{mpr:alp_lone} for all values of the regularization coefficient $\rc$,
it is possible to increase the coefficient $\rc$ until the error increase
between sampled constraints balances out the decrease in the error due to the
restricted feature space as defined in \aref{thm:regbound_offline}. That is,
we can calculate the objective value of the linear program
\eqref{mpr:alp_lone} for any value of $\rc$.

It is easy to find $\rc$ that minimizes the bounds in this section. As
the following corollary shows, to find the global minimum of the bounds, it is
sufficient to use the homotopy method to trace $\objregone(\rc)$ while its
derivative is less than the increase in the error due to the sampling ($\| \hat\val - \tilde\val \|_{1,\tobj}$). Let $\val(\rc)$ be an optimal
solution of \eqref{mpr:alp_estimated} as a function of the regularization
coefficient $\rc$.
\begin{cor} \label{cor:select_rc}
Assume that $\epsilon_c(\rc)$, $\epsilon_p(\rc)$, and $\epsilon_s(\rc)$ are
convex functions of $\rc$.  Then, the error bound $\| \val(\rc) - \val^*
\|_{1,\tobj} \leq f(\rc)$ for any $\val(\rc)$ is:
\[ f(\rc) = \objregone(\rc) - \tobj\tr \val^* + \epsilon_c(\rc) + 2\frac{\epsilon_s(\rc) + \epsilon_p(\rc)}{1-\disc}. \]
The function $f(\rc)$ is convex and its sub-differential\footnote{Function $f$
may be non-differentiable} $\nabla_{\rc} f$ is independent of $\val^*$.
Therefore, a global minimum $\rc^*$ of $f$ is attained when $0 \in
\nabla_{\rc} f(\rc^*)$ or when $\rc^*=0$.
\end{cor}
The corollary follows directly from \aref{thm:regbound_online} and the convexity of the optimal objective value of \eqref{mpr:alp_lone} as a function $\rc$. \aref{fig:sampling_bounds} illustrates the functions in the corollary. Notice that \aref{cor:linear_values} is sufficient to satisfy the conditions of this
corollary. In particular, the functions $\epsilon_s(\rc),\epsilon_p(\rc),\epsilon_c(\rc)$ are linear in $\rc$.

%% file: experimental.tex
\section{Experimental Results}
\label{sec:experimental}

In this section, we present results indicating that RALP effectively selects
from rich feature spaces to outperform ALP and other common algorithms, such
as LSPI, on several example problems, including the balanced pendulum and the
bicycle problems. We also demonstrate the speed and effectiveness of the
homotopy method in choosing a value of $\rc$.

\paragraph{Benefits of Regularization} First, we demonstrate and analyze the
properties of RALP on a simple chain problem with 200 states, in which the
transitions move to the right by one step with a centered Gaussian noise with
standard deviation 3. The reward for reaching the right-most state was +1 and
the reward in the 20th state was -3. This problem is small to enable
calculation of the optimal value function and to control sampling. We
uniformly selected every fourth state on the chain and estimated the sampling
bound $\epsilon_p(\rc) = 0.05\rc$. The approximation basis in this problem is
represented by piecewise linear features, of the form $\phi(s_i)=\pos{i-c}$,
for $c$ from 1 to 200; these features were chosen due to their strong
guarantees for the sampling bounds. The experimental results were obtained
using the proposed homotopy algorithm.

\begin{figure*}
\centering
\begin{minipage}{0.27\linewidth}
\centering
\includegraphics[width=\linewidth]{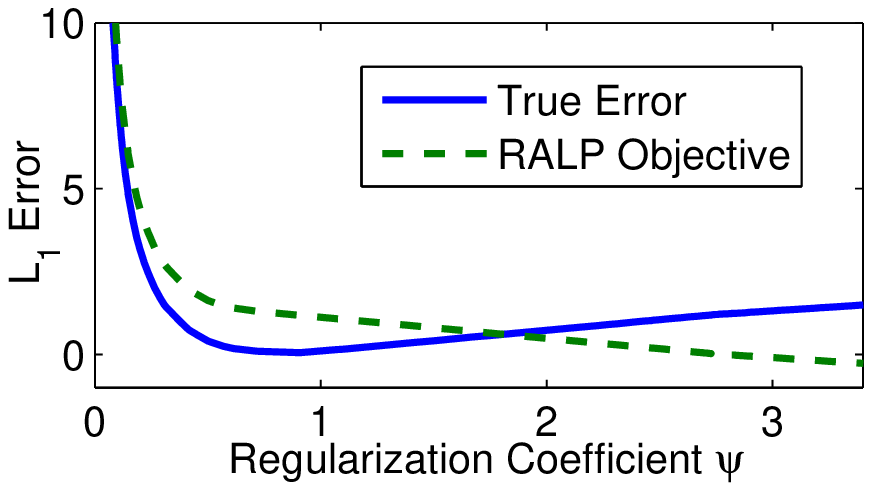}
\caption{Comparison of the objective value of RALP with the true error.} \label{fig:overfitting}
\end{minipage} \quad
\begin{minipage}{0.27\linewidth}
\centering
\includegraphics[width=\linewidth]{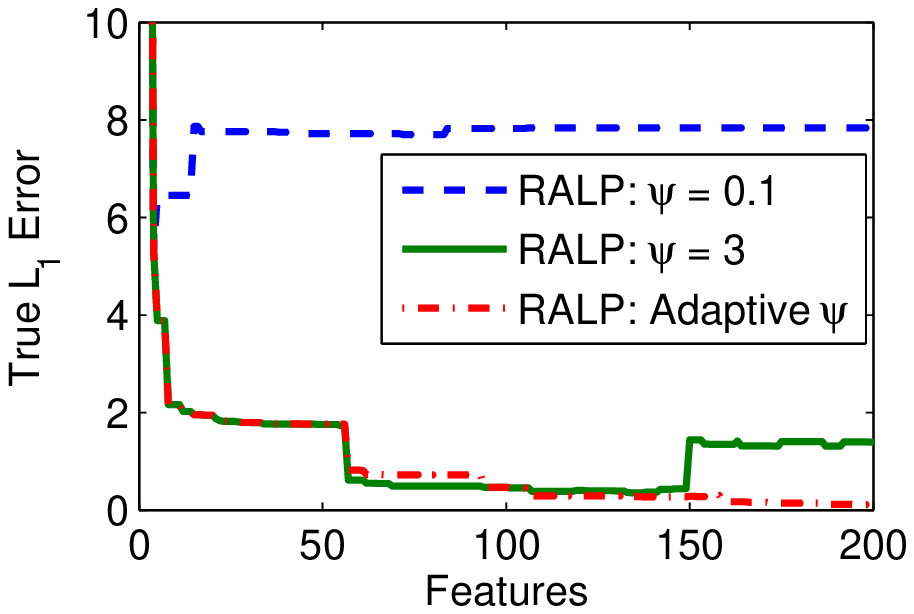}
\caption{Comparison of the performance RALP with two values of $\rc$ and the
one chosen adaptively (\aref{cor:select_rc}).} \label{fig:automatic}
\end{minipage} \quad
\begin{minipage}{0.27\linewidth}
\centering
\includegraphics[width=\linewidth]{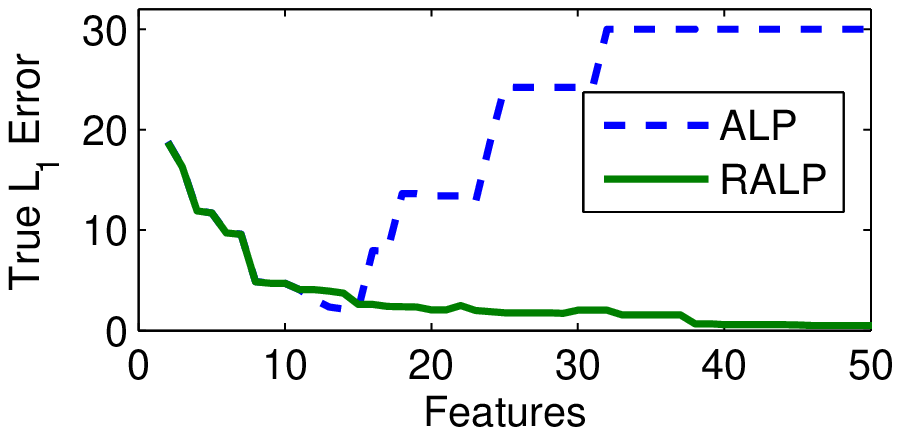}
\caption{Average of 45 runs of ALP and RALP as a function of the number of
features. Coefficient $\rc$ was selected using \aref{cor:select_rc}.}
\label{fig:feature_selection}
\end{minipage}
\end{figure*}

\aref{fig:overfitting} demonstrates the solution quality of RALP on the chain
problem as a function of the regularization coefficient $\rc$. The figure
shows that although the objective of RALP keeps decreasing as $\rc$ increases,
the sampling error overtakes that reduction. It is clear that a proper
selection of $\rc$ improves the quality of the resulting approximation. To
demonstrate the benefits of regularization as it relates to overfitting, we
compare the performance of ALP and RALP as a function of the number of
available features in \aref{fig:feature_selection}. While ALP performance
improves initially, it degrades severely with more features. The value $\rc$
in RALP is selected automatically using \aref{cor:select_rc} and the sampling
bound of $\epsilon_p(\rc) = 0.05\rc$. \aref{fig:automatic} demonstrates that RALP may
also overfit, or perform poorly when the regularization coefficient $\rc$ is
not selected properly.

\begin{figure*}
\centering
\begin{minipage}{0.27\linewidth}
\centering
\includegraphics[width=\linewidth]{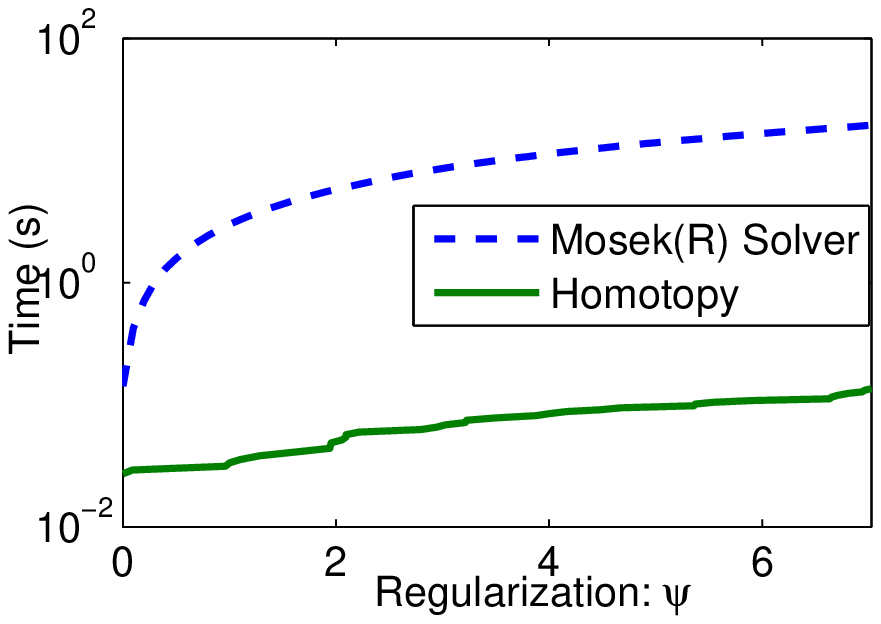}
\caption{Comparison of performance of homotopy method versus Mosek as a
function of $\rc$ in the mountain car domain.}
\label{fig:homovsrc}
\end{minipage}\quad
\begin{minipage}{0.27\linewidth}
\centering
\includegraphics[width=\linewidth]{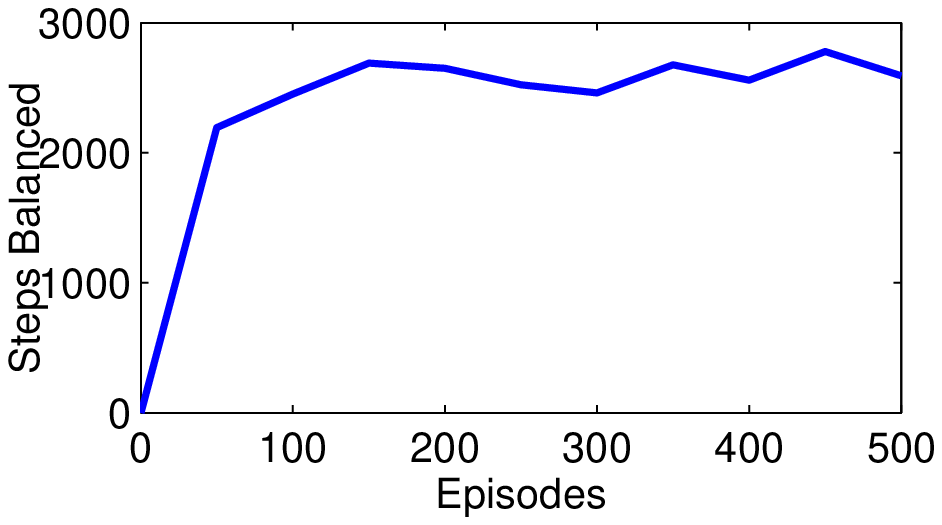}
\caption{RALP performance on pendulum as a function on the number of
episodes.}
\label{fig:pendulum}
\end{minipage} \quad
\begin{minipage}{0.27\linewidth}
\centering
\includegraphics[width=\linewidth]{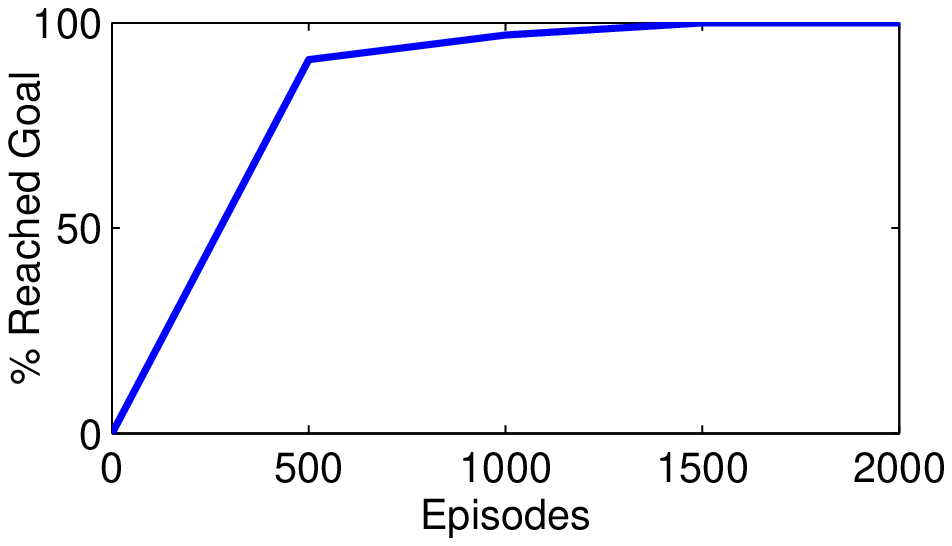}
\caption{RALP performance on bicycle as a function on the number of episodes.} \label{fig:bike}
\end{minipage}
\end{figure*}

To find the proper value of $\rc$, as described in \aref{cor:select_rc}, the
problem needs to be solved using the homotopy method. We show that the
homotopy method performs significantly faster than a commercially available
linear program solver Mosek. \aref{fig:homovsrc} compares the computational
time of homotopy method and Mosek, when solving the problem for multiple
values of $\rc$ in increments of 0.5 on the standard mountain car
problem~\cite{barto98} with 901 piecewise linear
features and 6000 samples. Even for any \emph{single} value $\rc$, the
homotopy method solves the linear program about 3 times faster than Mosek. The next two experiments, however, do \emph{not} use the homotopy method.  In
practice, RALP often works much better than what is suggested by our bounds,
which can be loose for sparsely sampled large state spaces.  In the following
experiments, we determined $\rc$ empirically by solving the RALP for several
different values of $\rc$ and selecting the one that produced the best policy.
This was practical because we could solve the large RALPs in just a few
minutes using constraint generation.


\paragraph{Inverted Pendulum} We now offer experimental results demonstrating
RALP's ability to create effective value functions in balancing an inverted
pendulum, a standard benchmark problem in reinforcement
learning~\cite{wang96,lspi}. Samples of the form $(s,a,r,s')$ were drawn from
every action on states drawn from random trajectories with the pendulum
starting in an upright state, referred to as episodes. Features were kernels
on 650 states randomly selected from the training data, consisting of Gaussian
kernels of standard deviation 0.5, 1, and 1.5, and a 6$^{th}$ degree polynomial
kernel.  $\rc$ was 1.4, and an average of 25 features had non-zero weights.


The constraints in the RALP were based on a single sample for each state and
action pair. The policy was evaluated based on the number of steps it could
balance the pendulum, with an upper limit of 3000. This served to evaluate the
policy resulting from the approximate value function.  We plot the average
number of steps the pendulum was balanced as a function of the number of
training episodes in \aref{fig:pendulum}, as an average of 100 runs. It is
clear the controller produced by RALP was effective for small amounts of data,
balancing the pendulum for the maximum number of steps nearly all of the time,
even with only 50 training episodes. Similarly, it was able to leverage the
larger number of available features to construct an effective controller with
fewer trajectories than LSPI, which needed 450 training episodes before
achieving an average of 2500 balanced steps~\cite{lspi}.

\paragraph{Bicycle Balancing and Riding} We also present experimental results
for the bicycle problem, in which the goal is to learn to balance and ride a
bicycle to a target position~\cite{randlov98,lspi}. This is a challenging
benchmark domain in reinforcement learning. Training data consisted of samples
for every action on states drawn from trajectories resulting from random
actions up to 35 states long, similar to the inverted pendulum domain. The
feature set consisted of monomials up to degree 4 on the individual dimensions
and products of monomials up to degree 3, for a total of 159 features.  $\rc$
was 0.03, and an average of 34 features had nonzero weights.  We plot the
number of runs out of 100 in which the bicycle reached the goal region as a
function of number of training episodes in \aref{fig:bike}.  Again, a high
percentage of runs were successful, even with only 500 training episodes.  In
comparison, LSPI required 1500 training episodes to pass 80\% success.  It is
worth pointing out that due to sampling every action at each state, RALP is
using more samples than LSPI, but far fewer trajectories.


%% file: conclusion.tex
\section{Conclusion and Related Work} \label{sec:conclusion}


In this paper, we introduced $\lone$-regularized Approximate Linear
Programming and demonstrated its properties for combined feature selection and
value function approximation in reinforcement learning.  RALP simultaneously
addresses the feature selection, value function approximation, and policy
determination problems; our experimental results demonstrate that it addresses
these issues effectively for several sample problems, while our bounds explain
the effects of sampling on the resulting approximation.

There are many additional issues that need to be addressed. The first is the
construction of better bounds to guide the sampling.  Our bounds explain the
behavior of RALP approximation as it relates to the trade-off between the
richness of the features with the number of available samples, but
these bounds may at times be quite loose.  Future work must identify
conditions that can provide stronger guarantees.  Additionally, a data-driven
approach which can calculate a tighter bound online would be valuable.
Finally, our analysis did not address the conditions that would guarantee
sparse RALP solutions and, therefore, allow more computationally efficient
solvers.

\subsection*{Acknowledgements}

This work was supported in part by DARPA CSSG HR0011-06-1-0027, by NSF
IIS-0713435, by the Air Force Office of Scientific Research under Grant No.
FA9550-08-1-0171, and by the Duke University Center for Theoretical and
Mathematical Sciences. We also thank the anonymous reviewers for their useful
comments.

%% file: proofs.tex
\section{Proofs} \label{sec:Proofs}
\subsection{Sampling Bounds} \label{sec:sampling_bounds}

The following lemmas summarize the basic properties of the Bellman operator.
We include them without proofs, which use standard techniques.

\begin{lem}[Bellman Operator] \label{lem:Bellman_one}
Let $\one$ and $\val$ be a constant vector of an appropriate size. Then:
\[\Bell (\val + \epsilon \one) = \disc \epsilon \one + \Bell \val.\]
\end{lem}

\begin{lem}[Transitive Feasible are Upper Bound] \label{lem:greater_transitive_feasible}
Assume an $\epsilon$-transitive-feasible value function $\val \in
\tf(\epsilon)$. That is:
\[\val \geq \Bell \val -  \epsilon \one. \]
Then:
\[ \val \geq \val^* - \frac{\epsilon}{1-\gamma} \one. \]
\end{lem}

\begin{lem} \label{lem:existence_infty_bound}
Assume \aref{asm:contains_one} and that there exists $\val \in \rep$ such that
\[\| \val - \val^*\|_\infty \leq \epsilon.\]
Then, there exists $\val' \in \rep \cap \tf$ such that
\[\| \val' - \val^* \|_\infty \leq \frac{2\epsilon}{1-\gamma}. \]
\end{lem}

\thmregboundoffline
\begin{proof}
To simplify the notation, we omit $\rc$ in the notation of $\epsilon$ in the proof.

\noindent Proof of $\| \hat \val - \val^* \|_{1,\tobj}$: \\
Let $\val$ be the minimizer of $\min_{\val\in\rep} \| \val - \val^*\|_\infty$. Then from \aref{lem:existence_infty_bound}, \aref{lem:greater_transitive_feasible}, and $\| x \|_{1,\tobj} \leq \| x \|_\infty$ there exists $\val' \in \rep \cap \tf$ such that:
\[ \| \val' - \val^* \|_\infty \leq \frac{2}{1-\disc} \| \val - \val^*\|_\infty \]
From \aref{lem:greater_transitive_feasible}, we have that:
\begin{align*} \| \hat\val - \val^* \|_{1,\tobj}
&= \tobj\tr (\hat\val - \val^*) = \tobj\tr\hat\val - \tobj\tr \val^* \leq \tobj\tr\val' - \tobj\tr \val^* \\
&\leq \| \val' - \val^* \|_{1,\tobj}.
\end{align*}

\noindent Proof of $\| \bar \val - \val^* \|_{1,\tobj}$:\\
Let $\bar\val'$ be the solution \eqref{mpr:alp_sampled} but with $\tobj\tr\val$
as the objective function. From \aref{lem:greater_transitive_feasible} we
have:
\begin{align*}
\bar\val &\geq \val^* - \frac{\epsilon_p}{1-\disc} \one \\
\bar\val' &\geq \val^* - \frac{\epsilon_p}{1-\disc} \one.
\end{align*}
The difference between $\bar\val$ and $\bar\val'$ can be quantified as
follows, using their same feasible sets and the fact that $\bar\val$ is minimal
with respect to $\bar{\tobj}$:
\[ \tobj\tr \bar\val \leq \bar{\tobj}\tr \bar\val + \epsilon_c \leq \bar{\tobj}\tr \bar\val' + \epsilon_c \leq \tobj\tr \bar\val' + 2\epsilon_c. \]
Since $\tf(\epsilon_p) \supseteq \tf$ we also have that $\tobj\tr \bar\val' \leq \tobj\tr \hat\val$. Then, using that $\hat\val \in \bar{\tf}$ and $\hat\val \geq \val^*$:
\begin{align*}
\| \bar\val - \val^* \|_{1,\tobj}
&\leq \left\| \bar\val - \val^* + \frac{\epsilon_p}{1-\disc} \one - \frac{\epsilon_p}{1-\disc} \one \right\|_{1,\tobj} \\
&\leq \left\| \bar\val - \val^* + \frac{\epsilon_p}{1-\disc} \one \right\|_{1,\tobj} + \frac{\epsilon_p}{1-\disc}  \\
&\leq \tobj\tr \left(\bar\val - \val^* + \frac{\epsilon_p}{1-\disc} \one\right) + \frac{\epsilon_p}{1-\disc} \\
&\leq \tobj\tr \left(\bar\val - \val^* \right) + 2 \frac{\epsilon_p}{1-\disc} \\
&\leq \tobj\tr \left(\bar\val' - \val^* \right) + 2\epsilon_c + 2 \frac{\epsilon_p}{1-\disc} \\
&\leq \tobj\tr \left(\hat\val - \val^* \right) + 2\epsilon_c + 2 \frac{\epsilon_p}{1-\disc} \\
&\leq \| \hat\val - \val^* \|_{1,\tobj} + 2 \epsilon_c + 2 \frac{\epsilon_p}{1-\disc}
\end{align*}

\noindent Proof of $\| \tilde \val - \val^* \|_{1,\tobj}$: \\
Let $\tilde\val'$ be the solution \eqref{mpr:alp_estimated} but with $\tobj$ as the objective function. Using \aref{lem:greater_transitive_feasible} we have:
\begin{align*}
\tilde\val &\geq \tilde{\Bell} \val \geq \bar{\Bell} \val - \epsilon_s \one \geq \Bell \val - (\epsilon_s + \epsilon_p) \one \\
\tilde\val &\geq \val^* - \frac{\epsilon_s + \epsilon_p}{1-\disc} \one
\end{align*}
All of these inequalities also hold for $\tilde\val'$ since it is also feasible in \eqref{mpr:alp_estimated}. From \aref{lem:Bellman_one}:
\[\hat\val + \frac{\epsilon_s}{1-\disc} \one \in \tilde{\tf}.\]
Therefore:
\[ \tobj\tr \tilde\val' \leq \tobj\tr \hat\val + \frac{\epsilon_s}{1-\disc}  \]
The difference between $\tilde\val$ and $\tilde\val'$ can be bounded as follows, using their same feasible sets and the fact that $\tilde\val$ is minimal with respect to $\bar{\tobj}$:
\[ \tobj\tr \tilde\val \leq \bar{\tobj}\tr \tilde\val + \epsilon_c \leq \bar{\tobj}\tr \tilde\val' + \epsilon_c \leq \tobj\tr \tilde\val' + 2\epsilon_c \]
Then:
\begin{align*}
\| \tilde\val - \val^* \|_{1,\tobj}
&\leq \left\| \tilde\val - \val^* + \frac{\epsilon_s + \epsilon_p}{1-\disc} \one - \frac{\epsilon_s + \epsilon_p}{1-\disc} \one \right\|_{1,\tobj} \\
&\leq \left\| \tilde\val - \val^* + \frac{\epsilon_s + \epsilon_p}{1-\disc} \one \right\|_{1,\tobj} + \frac{\epsilon_s + \epsilon_p}{1-\disc}  \\
&\leq \tobj\tr \left(\tilde\val - \val^* + \frac{\epsilon_s + \epsilon_p}{1-\disc} \one\right) + \frac{\epsilon_s + \epsilon_p}{1-\disc} \\
&\leq \tobj\tr \left(\tilde\val - \val^* \right) + 2 \frac{\epsilon_s + \epsilon_p}{1-\disc} \\
&\leq \tobj\tr \left(\tilde\val' - \val^* \right) + 2\epsilon_c + 2 \frac{\epsilon_s + \epsilon_p}{1-\disc} \\
&\leq \tobj\tr \left(\hat\val - \val^* \right) + 2\epsilon_c + \frac{3\epsilon_s + 2\epsilon_p}{1-\disc} \\
&\leq \| \hat\val - \val^* \|_{1,\tobj} + 2 \epsilon_c + \frac{3\epsilon_s + 2\epsilon_p}{1-\disc}
\end{align*}
\end{proof}

\thmregboundonline
\begin{proof}
To simplify the notation, we omit $\rc$ in the notation of $\epsilon$ in the proof. Using \aref{lem:greater_transitive_feasible} we have:
\begin{align*}
\tilde\val &\geq \tilde{\Bell} \val \geq \bar{\Bell} \val - \epsilon_s \one \geq \Bell \val - (\epsilon_s + \epsilon_p) \one \\
\tilde\val &\geq \val^* - \frac{\epsilon_s + \epsilon_p}{1-\disc} \one.
\end{align*}
Then, using the above:
\begin{align*}
\| \val - \val^* \|_{1,\tobj} &= \| \val^* - \val  + \frac{\epsilon_s + \epsilon_p}{1-\disc} \one - \frac{\epsilon_s + \epsilon_p}{1-\disc} \one \|_{1,\tobj} \\
&\leq \| \val - \val^*  + \frac{\epsilon_s + \epsilon_p}{1-\disc} \one \|_{1,\tobj} + \frac{\epsilon_s + \epsilon_p}{1-\disc} \\
&= \tobj\tr\val - \tobj\val^*  + 2 \frac{\epsilon_s + \epsilon_p}{1-\disc} \\
&= \bar{\tobj}\tr\val - \tobj\val^* + \epsilon_c + 2 \frac{\epsilon_s + \epsilon_p}{1-\disc}
\end{align*}
\end{proof}

\subsection{Sampling Guarantees} \label{sec:sampling_guarantees}

Let $k:\states\rightarrow\Real^n$ be a map of the state-space to an Euclidean
space. This basically means that the states of the MDP can be mapped to an
normed vector space. This assumption trivially generalizes
\aref{asm:lipschitz}.
\begin{asm} \label{asm:lipschitz_general}
Assume that samples $\bar{\samples}$ are available. Assume also that features and transitions satisfy Lipschitz-type constraints:
\begin{align*}
\| \feats(\bar{s}) - \feats(s) \|_{\infty} &\leq K_\phi \|k(s) - k(\bar{s})\|\\
| \rew(\bar{s}) - r(s) | &\leq K_r  \|k(s) - k(\bar{s})\| \\
\| \tran(\bar{s},a)\tr \feats_i - \tran(s,a)\tr \feats_i  \|_{\infty} &\leq K_p \|k(s) - k(\bar{s})\| \quad \forall a \in \actions
\end{align*}
Here $\feats_i$ denotes a vector representation of a feature across all states.
\end{asm}

It is easy to show the following:
\begin{prop}
Assume \aref{asm:lipschitz} and that $\| \feats_i \|_1 = 1$. Then \aref{asm:lipschitz_general} holds with the identical constants.
\end{prop}
The proof is simple and relies on the trivial version of Holder's inequality to combine $\| \tran(\bar{s},a)\tr \feats_i - \tran(s,a)\tr \feats_i \|_\infty$ with $\| \feats_i \|_1$.

While \aref{asm:lipschitz_general} characterizes general properties of the MDP, the following assumption unifies the assumptions on the MDP and the sampling.
\begin{asm}[Sufficient Sampling] \label{asm:sufficient_sampling}
Assume that for $\forall s \in \states$, there exists $\exists \bar{s} \in \bar{\states}$, such that:
\begin{itemize}
    \item $\|\feats(\bar{s}) - \feats(s) \|_{\infty} \leq \delta_\phi$
    \item $|\rew(\bar{s}) - \rew(s) | \leq \delta_r$
    \item $\|\tran(\bar{s},a)\tr \feats_i - \tran(s,a)\tr \feats_i  \|_{\infty} \leq \delta_p \quad \forall a \in \actions$
\end{itemize}
where $\tran(s,a)$ represents the vector of transition probabilities from
states $s$ given action $a$.
\end{asm}

\begin{thm} \label{thm:sampling_close}
Assume \aref{asm:contains_one} and \aref{asm:sufficient_sampling}. Let the value function $\val$ be represented as $\val \in\rep$ such that $\| x \|_1 \leq \rc$. Then \aref{asm:sampling_behavior} is satisfied with the following constraints:
\begin{align*}
    \epsilon_p &= \delta_r + \rc (\delta_\feats + \disc \delta_p ) \\
\end{align*}
The theorem also holds if a column of $\repm$ is $\one$ and the corresponding element of $x$ is not included in the norm.
\end{thm}
\begin{proof}
From the assumptions in the theorem, we have that $v \geq \bar{L} v$ and we need to show that $v \geq L v - \epsilon_p \one$. Then we get, using Holder's inequality that:
\begin{align*}
\min_{s\in\states}& (v - L v)(s)  \geq \\
    &\geq \min_{s\in\states}  (v - L v)(s) - (v - \bar{L}v)(\bar{s})  \\
    &\geq -\max_{s\in\states,a\in\actions} |((\feats(s) - \disc \tran(s,a)\tr\repm) \\
    &\quad - (\feats(\bar{s})-\disc \tran(\bar{s},a)\tr\repm ) x) -\rew(s) + \rew(\bar{s})| \\
    &\geq -\max_{s\in\states,a\in\actions} \|(\feats(s) - \disc \tran(s,a)\tr\repm) \\
    &\quad - (\feats(\bar{s})-\disc \tran(\bar{s},a)\tr\repm ) \|_\infty \|x\|_1 + |\rew(s) + \rew(\bar{s})| \\
    &\geq -\max_{s\in\states,a\in\actions} \|(\feats(s) - \disc \tran(s,a)\tr\repm) \\
    &\quad - (\feats(\bar{s})-\disc \tran(\bar{s},a)\tr\repm ) \|_\infty \rc + \delta_r \\
    &\geq -\max_{s\in\states,a\in\actions} \|(\feats(s) -\feats(s) \|_1 + \disc \| \tran(s,a)\tr\repm)\\
    &\quad  - (\feats(\bar{s})- \tran(\bar{s},a)\tr\repm ) \|_\infty \rc + \delta_r \\
    &\geq -(\delta_\feats +\delta_p)\rc - \delta_r
\end{align*}
\end{proof}

The following then summarizes the results.
\corlinearvalues
\begin{proof}
The corollary follows simply by setting the following values.
\begin{align*}
    \delta_\phi &= c K_\phi \\
    \delta_r &= c K_r\\
    \delta_p &= c K_P
\end{align*}
\end{proof}

Notice that for simplicity, we did not provide any bounds on $\epsilon_c$. These bounds require additional assumptions on the sampling procedure. That is the sampling must not only cover the whole space, but must also be uniformly distributed over it. Without such distribution, the objective function $\tobj$ must be a \emph{weighted} sum of the states.

%% file: homotopy.tex
\section{Homotopy Continuation Method} \label{sec:Homotopy}

The homotopy algorithm is similar to sensitivity analysis of the standard
simplex algorithm. However, the basic feasible solutions in simplex are of the
size of the number of variables. Because we are interested in solving very
large linear programs, this is often impractical. The homotopy method we
propose instead relies on basic basic feasible solutions with size that
corresponds to the number of variables.

As before, we use $\one_i$ to denote a zero vector with $i$-th element set to
1. For a matrix $A$, we use $A_j$ to denote the $j$-th row and $A_{\cdot i}$
as $i$-the column. We also use $x(i)$ to denote the $i$-the element of the
vector. We derive the algorithm for a generic linear program, defined as
follows:
\begin{mprog} \label{mpr:primal_lone}
\minimize{x} c\tr x
\stc A x \geq b
\cs \reg \tr x \leq \rc
\cs x \geq 0
\end{mprog}
Note that the variables are constrained to be non-negative. Any unbounded
variable $z$ can be expressed as $z = z_{+} - z_{-}$, such that $z_{+},z_{-}
\geq 0$. The homotopy algorithm also relies on the dual formulation of the
linear program \eqref{mpr:primal_lone}, which is as follows.
\begin{mprog}\label{mpr:dual_lone}
\maximize{y,\lambda} b\tr y - \rc \lambda
\stc A\tr y - \reg \lambda  \leq c
\cs y, \lambda \geq 0
\end{mprog}

The algorithm traces the optimal solution of the linear program
\eqref{mpr:primal_lone} as a function of $\rc$, which is defined as follows.
\begin{defn}
The optimal solution of the linear program \eqref{mpr:primal_lone} as a
function of $\rc$ is denoted as $x(\rc)$, assuming that the optimal solution
is a singleton. Notice that this is the optimal solution, not the optimal
objective value. We also use  $y(\rc)$ and $\lambda(\rc)$ similarly to denote
the sets of optimal solutions of the dual program \eqref{mpr:dual_lone} for
the given the regularization coefficient.
\end{defn}

The  homotopy algorithm keeps a set of active variables $\avars(x,y)$ and a
set of active constraints $\acons(x,y)$. A variable is considered to be active
if it is non-zero. A constraint is considered to be active when the
corresponding dual value is non-negative. Active and inactive variables and
constraints are formally defined as follows.
\begin{align*}
\avars(x,y) &= \{ i \ss x(i) \geq 0 \} &
\nvars(x,y) &= \{ i \ss x(i) = 0 \} \\
\acons(x,y) &= \{ j \ss y(j) \geq 0 \} &
\ncons(x,y) &= \{ j \ss y(j) = 0 \}
\end{align*}
We use $\avars$ in place of $\avars(x,y)$ when the values of $x,y$ are
apparent from the context. Notice that this definition is intentionally
ambiguous, that is for a given value of $x,y$, the active variables and
constraints are not uniquely specified. This is intentional, to allow for
adding and removing them at the points of discontinuity. Although, active
primal and dual variables may be 0, the inactive variables always must be 0.
The active variables and constraints can be used to define the following
variables.
\begin{align*} \label{eq:partition}
c &= \begin{pmatrix} c_{\avars} \\ c_{\nvars} \end{pmatrix} &
x &= \begin{pmatrix} x_{\avars} \\ x_{\nvars} \end{pmatrix} \\
b &= \begin{pmatrix} b_{\acons} \\ b_{\ncons} \end{pmatrix} &
y &= \begin{pmatrix} y_{\acons} \\ y_{\ncons} \end{pmatrix} &
A &= \begin{pmatrix} A_{\avars \acons} & A_{\nvars \acons} \\ A_{\avars \ncons} & A_{\nvars \ncons} \end{pmatrix}
\end{align*}
We assume the given order of variables.

The following assumptions are needed in order to derive the algorithm.
\begin{asm}
The optimal solution of \eqref{mpr:primal_lone} is feasible and bounded for
values of $\rc \in [0,\infty)$. In addition, it is ``easy'' to solve for
$\rc=0$.
\end{asm}
\begin{asm} \label{asm:at_most_one}
For any $\rc \in [0,\infty)$ the solution of \eqref{mpr:primal_lone} is not
degenerate.
\end{asm}
These assumptions guarantee that at no point the solutions become degenerate.
This is a common assumption in simplex algorithms, and can be remedied for
example by assuming a small perturbation of variables~\cite{Vanderbei2001}.

The homotopy algorithm traces the optimal solution of the linear program,
which can be characterized by a set of linear equations. These optimality
conditions for \eqref{mpr:primal_lone} can be derived identically from KKT or
from the complementary slackness optimality conditions as follows.
\begin{align*}
-\reg\tr x &\geq - \rc \\
A x  &\geq b \\
A\tr y &\leq c + \lambda \reg  \\
y\tr (b - A x ) &= 0 \\
\lambda (\reg\tr - \rc) &= 0 \\
x\tr ( c - A\tr y + \reg \lambda  ) &= 0 \\
x,y,\lambda &\geq 0
\end{align*}
Without loss of generality, we assume that the active constraints and
variables are the first in the respective structures. This does not impose any
limitations, and could be expressed generally using a permutation matrix $P$.
We implicitly assume that the regularization vector $\reg$ is partitioned
properly for the active variables.

For a given set of active variables and constraints, and using the fact that
the inactive variables $x$ and $y$ are 0, the optimality conditions can be
then rewritten as:
\begin{align*}
\reg\tr x_{\avars} &= \rc  \\
A_{\avars \acons} x_{\avars} &= b_{\acons} & A_{\avars \ncons} x_{\avars} &\leq b_{\ncons} \\
A_{\avars \acons}\tr y_{\acons} &= c_{\avars} + \lambda \reg & A_{\nvars \acons}\tr y_{\acons} &\leq 0 \\
x,y,\lambda &\geq 0
\end{align*}
We are assuming here that the regularization constraint is active. If it
becomes inactive at $\bar{\rc}$, the solution is optimal for any value of $\rc
\geq \bar{\rc}$. The equalities follow from the complementarity conditions,
omitted here. Notice that other constraints may also hold with equality.

\begin{algorithm*} \label{alg:homotopy}
$\rc^0 \leftarrow 0$ \;
\tcp{Find an initial feasible solutions}
$x^0 \leftarrow  x(\rc^0)$ \;
$y^0 \leftarrow y(\rc^0)$ \;
\tcp{Determine the initial active sets, and set $\nvars$ and $\ncons$ to be
their complements}
$\avars^0 = \{ i \ss x^0 > 0 \} $
$\acons^0 \leftarrow \{ j \ss y(j) > 0 \}$
\tcp{The regularization constraint is active, or the solution is optimal}
\While{$\rc^i < \bar{\rc}$ and $\lambda^i > 0 $}{
    $i \leftarrow i + 1$ \;
    \tcp{Here $|\acons| = |\avars| + 2$}
    \tcp{Calculate the space (line) of dual solutions -- the update direction}
    $\begin{pmatrix} \Delta y^i \\ \Delta \lambda^i\end{pmatrix} \leftarrow \nullspace \begin{pmatrix} A_{\avars \acons}\tr & \reg \end{pmatrix}$ such that $y^{i-1}(\rc) = 0 \Rightarrow \Delta y^i(\rc) \geq 0$ \tcp*{This is always possible because there is always at most one such constraint, given the assumptions.}
    \tcp{Decide based on a potential variable improvement}
    \tcp{Calculate the maximum length of the update $\tau$, breaking ties arbitrarily. }
    $t_1 \leftarrow \invm{ \max_{k \in \acons} \frac{-\Delta y_k^i}{y^{i-1}_k}}$ \tcp*{Some $y$ becomes 0.}
    $t_2 \leftarrow \invm{\max_{k \in \nvars} \frac{- \left(A_{\nvars \acons} \Delta y^i\right)_k}{\left(A_{\nvars \acons} y^{i-1} \right)_k}}$ \tcp*{Some $x$ needs to be added to the active set.}
    $t_3 \leftarrow \frac{\lambda}{-\Delta \lambda} $ \tcp*{Regularization constraint }
    $\tau = \min \left\{t_1, t_2, t_3 \right\}$
    \tcp{Resolve the non-linearity update, where $K_l$ is the set of maximizers for $t_l$ }
    \If{$\tau = t_1$}{ $\acons^i \leftarrow \acons^{i-1} \setminus K_1,\quad \ncons^i \leftarrow (\acons^i)\cmp$ }
    \ElseIf{$\tau = t_2$}{$\avars^i \leftarrow \avars^{i-1} \cup K_2,\quad \nvars^i \leftarrow (\avars^i)\cmp$}
    \ElseIf{$\tau = t_3$}{The regularization constraint is inactive, return the solution.}
    \tcp{Update the \emph{dual} solutions}
    $y^i \leftarrow y^{i-1} + \tau \Delta y^i ,\quad \lambda^i \leftarrow \lambda^{i-1} + \tau \Delta \lambda^i$ \;
    \tcp{Here $|\acons| = |\avars| + 1$}
    \tcp{Calculate the update direction}
    $\Delta x \leftarrow \inv{\begin{pmatrix} A_{\avars \acons} \\ \reg_{\avars} \end{pmatrix}} \begin{pmatrix} \zero \\ \Delta \rc \end{pmatrix}$ \;
    \tcp{Calculate the maximum length of the update $\tau$, breaking ties arbitrarily. }
    $t_4 \leftarrow \invm{\max_{k \in \ncons} \frac{-a_k\tr \Delta x^i}{a_k\tr x^{i-1} - b}}$ \tcp*{A constraint becomes active}
    $t_5 \leftarrow \invm{\max_{k \in \avars} \frac{-\Delta x^i_k}{a_k\tr x^{i-1} - b}}$ \tcp*{A variable
    $\tau = \min \left\{t_4, t_5 \right\}$}
   \tcp{Update the \emph{primal} solutions}
    $x^i \leftarrow x_{i-1} + \tau \Delta x^i$ \;
    \tcp{Resolve the non-linearity update, where $K_l$ is the set of maximizers for $t_l$ }
    \If{$\tau = t_4$}{$\acons^i \leftarrow \acons^{i-1} \cup K_4,\quad \ncons^i \leftarrow (\acons^i)\cmp$}
    \ElseIf{$\tau = t_5$}{$\avars^i \leftarrow \avars^{i-1} \setminus K_5,\quad \nvars^i \leftarrow (\avars^i)\cmp$}
}
\caption{Homotopy Continuation Method for Solving ALP} 
\end{algorithm*}

The homotopy algorithm is included in \aref{alg:homotopy}. The primal update
of the algorithm traces the solution in the linear segments and the dual
updates determines the update direction the sections with non-linearities. The
algorithm implementation is written with the focus on simplicity; an efficient
implementation relies on factorization of the matrices.
Two homotopy continuation methods --- DASSO~\cite{James2009} and primal dual
pursuit \cite{Asif2009} --- have been proposed for the Dantzig selector. These
homotopy methods are very efficient when the problems have sparse solutions,
as is often the case with Dantzig selectors. RALP cannot be solved directly
using the methods for the Dantzig selector, however, because of its different
structure. This structure can be used to develop a different homotopy method
for solving RALP, which is described in \aref{sec:Homotopy}. In addition, a
method based on parametric linear program solvers has been developed for
regularized linear programs.  However, the parametric simplex algorithm cannot
take advantage of sparse RALP solutions and therefore is not applicable to
large RALP. The finite-time convergence of \aref{alg:homotopy} can be,
however, shown identically to DASSO, or other related algorithms.